# TOWARDS A GENERAL-PURPOSE BELIEF MAINTENANCE SYSTEM

Brian Falkenhainer
Qualitative Reasoning Group
Department of Computer Science
University of Illinois
1304 W. Springfield Ave
Urbana, Illinois, 61801

## ABSTRACT

This paper addresses the problem of probabilistic reasoning as it applies to Truth Maintenance Systems. A *Belief Maintenance System* has been constructed which manages a current set of probabilistic beliefs in much the same way that a TMS manages a set of true/false beliefs. Such a system may be thought of as a generalization of a Truth Maintenance System. It enables one to reason using normal two or three-valued logic or using probabilistic values to represent partial belief. The design of the Belief Maintenance System is described and some problems are discussed which require further research. Finally, some examples are presented which show the utility of such a system.

## 1. INTRODUCTION

Truth Maintenance Systems have been used extensively in problem solving tasks to help organize a set of facts and detect inconsistencies in the believed state of the world. These systems maintain a set of true/false propositions and their associated dependencies. In trying to reason about real world problems, however, situations often arise in which we are unsure of certain facts or in which the conclusions we can draw from available information are somewhat uncertain. The non-monotonic TMS (Doyle, 1979) was an attempt at reasoning when we don't know all the facts. Non-monotonic systems, however, fail to take into account degrees of belief and how available evidence can combine to strengthen a particular belief.

This paper addresses the problem of probabilistic reasoning as it applies to Truth Maintenance Systems. It describes a *Belief Maintenance System* that manages a current set of beliefs in much the same way that a TMS manages a current set of true/false propositions. If the system knows that belief in $fact_1$ is dependent in some way upon belief in $fact_2$, then it automatically modifies belief in $fact_1$ if we give it some new information which causes a change in belief of $fact_2$. It models the behavior of a normal TMS, replacing its 3-valued logic (true, false, unknown) with an infinite-valued logic, in such a way as to reduce to a standard TMS if all statements are given in absolute true/false terms. Belief Maintenance Systems can, therefore, be thought of as a generalization of Truth Maintenance Systems, whose possible reasoning tasks are a superset of those for a TMS.

## 2. DESIGN

The design of the belief maintenance system is based on current TMS technology, specifically a monotonic version of Doyle's justification-based TMS (1979). As in the TMS, a network is constructed which consists of nodes representing facts and justification links between nodes representing antecedent support of a set of nodes for some consequent node. The BMS differs in that nodes take on a *measure of belief* rather than true or false and justification links become *support links* in that they provide partial evidence in favor of a node.

The basic design consists of three parts: (1) the conceptual control structure, (2) the user hooks to the knowledge base, and (3) the belief formalism. A simple parser is used to translate user assertions (e.g. `(implies (and a b) c)`) into control primitives. This enables the basic design to be semi-independent of the belief system used. All that is required of the belief formalism is that it is invertible. Specifically, if A provides support for B and our belief in A changes, we must be able to remove the effects the previous belief in A had on our belief in B.

### 2.1. Overview of Dempster-Shafer Theory

The particular belief system used here is based on the Dempster-Shafer theory of evidence (Shafer, 1976; Barnett, 1981; Garvey et al, 1981). Shafer's representation expresses the belief in some proposition A by the interval $|s(A), p(A)|$, where $s(A)$ represents the current amount of support for A and $p(A)$ is the plausibility of A. It is often best to think of $p(A)$ in terms of the lack of evidence against A, for $p(A) = 1 - s(\neg A)$. To simplify calculations, the BMS represents Shafer intervals by the pair $(s(A)\ s(\neg A))$ rather than the interval $|s(A)\ p(A)|$ (Ginsberg, 1984).

Dempster's rule provides a means for combining probabilities based upon different sources of information. His language of belief functions defines a *frame of discernment*, $\Theta$, as the exhaustive set of possibilities or values in some domain. For example, if the domain represents the values achieved from rolling a die, $\Theta$ is the set of 6 propositions of the form "the die rolled a j." If $m_1$ and $m_2$ are two basic probability functions over the same space $\Theta$, each representing a different knowledge source, then Dempster's *orthogonal sum* defines a new combined probability function $m$ which is stated as $m = m_1 \oplus m_2$.

Since the primary concern here is the use of probability theory in a deductive reasoning system, we are interested in the case where $\Theta$ contains only two values, A and $\neg A$. For this case, the basic probability function has only three values, $m(A)$, $m(\neg A)$, and $m(\Theta)$. This allows the derivation of a simplified version of Dempster's formula (Prade, 1983; Ginsberg, 1984):

$$(a\ b) \oplus (c\ d) = \left(1 - \frac{\bar{a}\bar{c}}{1 - (ad + bc)}\quad 1 - \frac{\bar{b}\bar{d}}{1 - (ad + bc)}\right)$$

where $\bar{a}$ means $(1 - a)$. It also allows us to formulate an inverse function for subtracting evidence (Ginsberg, 1984):

$$(a\ b) - (c\ d) = \left(\frac{\bar{c}(a\bar{d} - \bar{b}c)}{\bar{c}\bar{d} - bc\bar{c} - \bar{a}dd}\quad \frac{\bar{d}(b\bar{c} - \bar{a}d)}{\bar{c}\bar{d} - bc\bar{c} - \bar{a}dd}\right)$$

The decision to choose Dempster-Shafer Theory over other systems of belief was purely pragmatic. Dempster-Shafer has been shown to be invertible, it distinguishes between absolutely unknown (no evidence) and uncertain, and it is simple to use. However, the design of the BMS is not based on a particular belief formalism and there should be little difficulty (as far as the BMS itself is concerned) in adapting it to use some other belief system.

## 2.2. A Logic of Beliefs

The conventional meaning of two-valued logic must be redefined in terms of evidence so that the system can interpret and maintain its set of beliefs based on the user-supplied axioms.

### NOT

Because Dempster-Shafer theory allows us to express belief for and belief against in a single probability interval, (not A) and A can simply be stored as the same proposition, A, where

$$(\text{not } A)_{(s, \neg s)} = A_{(\neg s, s)}$$

### AND

There are a number of approaches to the meaning of AND. The interpretation used here corresponds to that of (Garvey et al, 1981):

$$A_{(s_A, s_{\neg A})}$$
$$B_{(s_B, s_{\neg B})}$$
$$C_{(s_C, s_{\neg C})}$$
$$\overline{\rule{3cm}{0.4pt}}$$
$$(A \& B \& C)_{(\max(0, s_A + s_B + s_C - 2), \max(s_{\neg A}, s_{\neg B}, s_{\neg C}))}$$

The $-2$ term represents $(1 - \text{cardinality}(\text{conjuncts}))$.

### OR

The belief in OR is the maximum of the individual beliefs (Garvey et al, 1981):

$$A_{(s_A, s_{\neg A})}$$
$$B_{(s_B, s_{\neg B})}$$
$$\overline{\rule{3cm}{0.4pt}}$$
$$(A \vee B)_{(\max(s_A, s_B), \max(0, s_{\neg A} + s_{\neg B} - 1))}$$

### IMPLIES

There are two theories in the literature for the interpretation of IMPLIES using Dempster-Shafer. (Dubois and Prade, 1985) suggests that, for $A \rightarrow B$, we take into account the value of $\text{Bel}(B \rightarrow A)$. Because the BMS should be simple to use and because $\text{Bel}(B \rightarrow A)$ can be difficult to obtain, the use of IMPLIES will be the same as (Ginsberg, 1984; Dubois & Prade, 1985):

$$(A \rightarrow B)_{(s_I, s_{\neg I})}$$
$$A_{(s_A, s_{\neg A})}$$
$$\overline{\rule{3cm}{0.4pt}}$$
$$B_{(s_A s_I, s_A s_{\neg I})}$$

This adheres to the idea that if full belief in A implies $B_{0.8}$, then a half belief in A should imply $B_{0.4}$.

With these operators defined, the system can parse all user assertions and construct the necessary support links with the appropriate belief functions attached to them.

## 2.3. Support Links

A support link consists of a list of antecedent nodes, a consequent node, its current positive and negative support for its consequent, and a function for recalculating its support based on the current belief of the antecedents (when the support is provided by the user, forming a *premise link*, no such function exists). Figure 1 shows a sample support link network. The system recognises two types of support links – *hard links* and *invertible links*.

### 2.3.1. Hard Support Links

A hard support link is one which provides an absolute statement of its consequent's belief. For example, statements of the form
$$(\text{implies } x \ (\text{and } y \ z))$$
are translated into
$$(\text{implies } x \ y)$$
$$(\text{implies } x \ z)$$
As a result, nodes are never allowed to give support directly to an "and" node and the only support entering an "and" node must come from the individual conjuncts. A support link for an "and" node is therefore given the status "hard link" and the value of the consequent node equals the link's support. In Figure 1, if the belief in A changes, a new value is calculated for the conjunctive link using its attached formula for AND, and the node for (AND A B) is set to the new value.

### 2.3.2. Invertible Support Links

Links representing implication or user support act as only one source of evidence for their consequent node. Such links are designated invertible since a change in their support means that their old support must be subtracted (using the inverted form of Dempster's rule) before the new value is added. In Figure 1, if the belief in D changes, then the current support provided by D's link into C is subtracted, the link support is recalculated, and the new support is added to C (using Dempster's rule).

## 2.4. Control

The basic control structure of the BMS is similar to that of a TMS. When the belief in a node is modified, the affects of this new belief are propagated throughout the system. This is done by following the node's outgoing links and performing the appropriate operations for modifying hard and invertible links' support. Propagation of evidence may be defined so as to terminate early (Ginsberg, 1985a). If the change made to a node's belief state is sufficiently small, there is no need to propagate this change to every node dependent upon it. A threshold value, *propagation-delta*, is defined so that, when the change to a node's positive and negative beliefs are less than the threshold, the system will not continue to propagate changes past this node. The default threshold is $10^{-3}$.

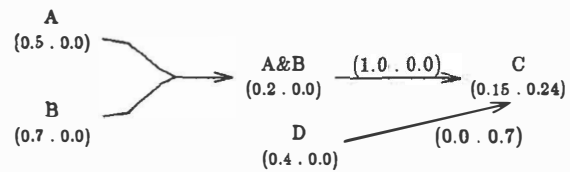

Figure 1. A Sample BMS Network

In a TMS architecture, only one justification is needed to establish truth. Any independent justifications are extraneous. Using a probabilistic architecture, each source of support adds to the node's overall belief. All incoming supports must be combined (using Dempster's rule) to form the overall belief for the node. If one tries to combine two contradicting, absolute beliefs, (1 0) $\oplus$ (0 1), the system would simply detect the attempt and signal a contradiction in the same way that a TMS would. Thresholds could also be used so that if a strongly positive belief is to be combined with a strongly negative belief, the system could signal a contradiction. Caution should be used for this case, however, because non-monotonic inferences should not be interpreted as contradictions.

### 2.4.1. New Control Issues

Circular support structures like that of Figure 2 cause a number of problems for belief maintenance. Because of these problems, the current implementation requires that no such structures exist and it will signal an error if one is discovered. There are a variety of problems which the structure in Figure 2 can cause:

(1) *Interpretation of circular evidence.* When A is partially believed and the status of E is unknown, what can be said about the support which D provides to B? All of the evidence D is supplying to B originally came from B in the first place. Because all links entering B will combine according to Dempster's rule to form a single belief, B may be believed more strongly than A simply because B supplies evidence in favor of D through C. This does not seem intuitively correct.

(2) *Problems with possible cures.* There are several potential solutions to this problem. First, D could be allowed to provide support for B. This situation is undefined under normal probability theory. Second, the chain can be stopped at D by not allowing any node to provide support to one of its supporters (by transitivity). This introduces a new problem. What should happen when E is providing independent support for D? Forcing the system to only propagate those supports for D which are independent of B would require a much more sophisticated control structure.

(3) *Retraction or modification of support.* Modifying support links becomes much more difficult if circular support structures are allowed to exist in the system. Any time the support A provides for B changes, the old support A provided must be retracted. This means removing all support from A, propagating the change in B, adding in the new support from A, and propagating the new belief in B. This will cause the belief in C to be propagated four times (twice when B changes the first time and twice when B changes the second time), the belief in D to be propagated 8 times, etc. In addition, retracting the support A provides for B means that we must retract all support for B (to remove the effects D has on B), propagate the new lack of belief in B, and then recalculate a new belief for B based on the new value for A and the current values of its other support links. Doing this every time the belief for any node changes makes such a system untenable. When we assume there is no circular support in the network, modifying belief in A simply involves subtracting its old support for B, adding in its new support for B, and then propagating the new belief in B.

The use of beliefs also causes problems for systems that explicitly calculate transitivity relations. Suppose we were to assert A $\rightarrow$ C based on the knowledge A $\rightarrow$ B and B $\rightarrow$ C. This action would cause the system's belief in C to increase, even though we were simply making explicit information which already existed.

## 2.5. User Support

The system has been designed so that it will appear to operate in exactly the same manner as the standard justification-based TMS. Thus, it is able to handle assertions using the connectives AND, OR, NOT, and IMPLIES. If a contradiction occurs, the system will notify the user and seek to resolve the contradiction. In addition to the normal TMS operations, the BMS supports additional operations corresponding to its belief-oriented knowledge.

### 2.5.1. Queries

In the TMS, queries are of the form (true? *statement*). Since truth is measured here in terms of belief, the query language can be extended. Truth is redefined in terms of a threshold, so that a belief over a certain threshold is considered to be true.

```
true?                = belief+(node) > *belief-threshold*
false?               = belief-(node) > *belief-threshold*
unknown?             = belief+(node) < *belief-threshold*
                       and belief-(node) < *belief-threshold*
absolutely-true?     = belief+(node) = 1.0
absolutely-false?    = belief-(node) = 1.0
absolutely-unknown?  = belief+(node) = 0.0
                       and belief-(node) = 0.0
support-for          = belief+(node)
support-against      = belief-(node)
possible-true        = 1 - belief+(node)
possible-false       = 1 - belief-(node)
belief-uncertainty   = 1 - belief-(node) - belief+(node)
```

### 2.5.2. Frames of Discernment

In addition to the default usage of the simplified version of Dempster's rule, where each node is treated as a frame of discernment, $\Theta$, containing $\{A, \neg A\}$, the user may define a

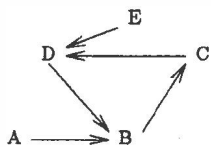

Figure 2. Circular Support Structure

73

specific frame of discernment by the function call:

(frame-of-discernment node$_1$ node$_2$ ... node$_n$)

This establishes a frame-of-discernment stating that the given nodes represent an exhaustive set of possibilities or values in some domain. Evidence in favor of one node acts to discredit belief in the other members of the set. Evidence may be provided to support any of the nodes from outside the set, but no support link is allowed to change once it has a non-zero value. This is due to the (current) uninvertibility of the general form of Dempster's rule. When new evidence is provided for one of the nodes in the set, the belief in all the nodes is recalculated according to Dempster's orthogonal sum so that the sum of the beliefs for the nodes in the set is less-than or equal-to one. The affect of these changes are then propagated to any support these nodes provide to the rest of the system.

### 2.6. Rule Engine

Because the BMS does not allow variables to exist in the knowledge base, pattern-directed rules are required to provide demons which trigger on certain events in the knowledge base (McAllester, 1980; Charniak et al, 1980). The rules are of the form:

(rule (nested-triggers) body)

For example, the rule

(rule ((:INTERN (dog ?x)))
    (assert (implies (dog ?x) (mammal ?x))))

causes the implication (implies (dog fido) (mammal fido)) to be asserted when (dog fido) first appears in the knowledge base (whether it is believed or not).

The rule

(rule ((:BELIEF+ (foo ?x) 0.8 :test (numberp ?x))
       (:BELIEF- (bar ?y) 0.9))
    (print "Support for Foo is now > 0.8")
    (print "Support against Bar is now > 0.9"))

fires when the belief in some (foo ?x), where ?x is a number, exceeds 0.8 and the belief in some (not (bar ?y)) exceeds 0.9. There are three types of rule triggers. The :INTERN trigger causes the rule to fire each time a new fact is added to the knowledge base which matches the given pattern. The BELIEF+ trigger causes the rule to fire each time the support in favor of an instance of its pattern first exceeds the specified value. A BELIEF- rule fires when the support against its pattern exceeds the specified value.

### 3. EXAMPLES

There are a number of possible uses for a belief maintenance system. It enables us to perform normal TMS, three-valued, deductive logic operations. For example, we can make assertions such as

(assert (implies a b))
(assert (implies b c))
(assert a)

and the system will automatically propagate the fact that b and c are true. If we then stated that c was false, the system would signal a contradiction and indicate that the contradiction results from the two user premises a and (not c).

The BMS is also able to reason with probabilistic or uncertain information. It is able to state the current partial belief in a particular item and the sources of this belief. In addition, a belief maintenance system is able to handle non-monotonic reasoning much more elegantly than a two or three valued logic is able to (Ginsberg, 1984, 1985b). Consider the classic non-monotonic problem about birds "in general" being able to fly. If one were to replace a rule about birds using Doyle's (1979) consistency operator with a probabilistic one stating that roughly 90 to 95% of all birds fly

$$(bird\ ?x) \rightarrow (fly\ ?x)_{(0.90\ 0.05)}$$

the desired non-monotonic behavior comes automatically from negative rules such as

$$(ostrich\ ?x) \rightarrow (fly\ ?x)_{(0\ 1)}$$

No modifications of the control structure are needed to perform non-monotonic reasoning.

### 3.1. Rule-based Pattern Matching

The belief maintenance system has been used to implement a rule-based, probabilistic pattern matching algorithm which is able to form the type of matching typical in analogies in a manner consistent with Gentner's Structure-Mapping Theory of analogy (Gentner, 1983; Falkenhainer, Forbus, & Gentner, 1986). For example, suppose we tried to match

(a) (AND (CAUSE (> (PRESSURE beaker) (PRESSURE vial))
         (FLOW beaker vial water pipe))
         (> (DIAMETER beaker) (DIAMETER vial)))

with

(b) (AND (> (TEMP coffee) (TEMP ice-cube))
         (FLOW coffee ice-cube heat bar))

A standard unifier would not be able to form the correspondences necessary for those two forms to match. First, the forms are different in their overall structure. Second, the arguments of similar substructures differ, as in (FLOW beaker vial water pipe) and (FLOW coffee ice-cube heat bar). The rule-based pattern matcher, however, is able to find all consistent matches between form (a) and form (b). These matches correspond to the possible interpretations of the potential analogy between (a) and (b). They are

(1)  (> (PRESSURE beaker) (PRESSURE vial))
     ↔ (> (TEMP coffee) (TEMP ice-cube))
     (FLOW beaker vial water pipe)
     ↔ (FLOW coffee ice-cube heat bar)

(2)  (> (DIAMETER beaker) (DIAMETER vial))
     ↔ (> (TEMP coffee) (TEMP ice-cube))

The pattern matcher works by first asserting a *match hypothesis* for each potential predicate or object pairing between (a) and (b) with a belief of zero. For example, we could cause all predicates having the same name to pair up and all functional predicates (e.g. PRESSURE) to pair up if their parent predicates pair up (e.g. GREATER). The likelyhood of each match hypothesis is then found by running *match hypothesis evidence rules*. For example, the rule

(assert same-functor)
(rule ((:intern (MH ?i1 ?i2)
        :test (and (fact? ?i1) (fact? ?i2)
                   (equal-functors? ?i1 ?i2))))
    (assert (implies same-functor (MH ?i1 ?i2)
            (0 5   0 0))))

| Match Hypothesis | Evidence |
|---|---|
| (MH GREATER$_{Pressure}$ GREATER$_{Temperature}$) | 0.650 |
| (MH GREATER$_{Diameter}$ GREATER$_{Temperature}$) | 0.650 |
| (MH PRESSURE$_{beaker}$ TEMPERATURE$_{coffee}$) | 0.712 |
| (MH PRESSURE$_{vial}$ TEMPERATURE$_{ice-cube}$) | 0.712 |
| (MH DIAMETER$_{beaker}$ TEMPERATURE$_{coffee}$) | 0.712 |
| (MH DIAMETER$_{vial}$ TEMPERATURE$_{ice-cube}$) | 0.712 |
| (MH FLOW$_{water}$ FLOW$_{heat}$) | 0.790 |
| (MH beaker coffee) | 0.932 |
| (MH vial ice-cube) | 0.932 |
| (MH water heat) | 0.632 |
| (MH pipe bar) | 0.632 |

Figure 3. BMS State After Running Match Hypothesis Evidence Rules

states "If the two items are facts and their functors are the same, then supply 0.5 evidence in favor of the match hypothesis." After running these rules, the BMS would have the beliefs shown in Figure 3.

The pattern matcher then constructs all consistent sets of matches to form *global matches* such that no item in a global match is paired up with more than one other item. (1) and (2) are examples of such global matches. Once the global matches are formed, the pattern matcher must select the "best" match. To do this, a frame of discernment consisting of the set of global matches is created and global match evidence rules are used to provide support for a global match based on various syntactic aspects such as overall size or "match quality". For example, we could have match hypotheses provide support in favor of the global matches they are members of. Thus, the pattern matcher would choose global match (1) because the match hypotheses provide the most support for this interpretation. This is a sparse description of the matching algorithm discussed in (Falkenhainer et al, 1986).

## 4. CONCLUSIONS

The design of a *belief maintenance system* has been presented and some of its possible uses described. This system differs from other probabilistic reasoning systems in that it allows dynamic modification of the structure of the knowledge base and maintains a current belief for every known fact. Previous systems have used static networks (Pearl, 1983; Buchanan et al, 1984) which cannot be dynamically modified or simple forward chaining techniques which don't provide a complete set of reason-maintenance facilities (Buchanan et al, 1984; Ginsberg, 1984, 1985).

There are still a number of unsolved problems. First, the interpretation and efficient implementation of circular support structures needs to be examined further. Second, operations such as generating explicit transitivity relations cause new problems for belief based reasoning systems. What is important to note is that the basic design is independent of the belief system used. For any given logic of beliefs (probabilities) which is invertible, the assertion parser can be modified to construct the appropriate network.

## 5. ACKNOWLEDGEMENTS

Thanks go to Peter Haddawy, Boi Faltings, Larry Rendell, Ken Forbus, and Barry Smith for their helpful discussions and proofreading skills. The BMS was modeled after a simple justification-based TMS written by Ken Forbus. This research is supported by the Office of Naval Research, Contract No. N00014-85-K-0559.

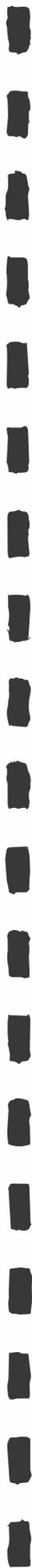